\documentclass[preprint,authoryear,12pt]{elsarticle}

\usepackage{amssymb}
\usepackage{amsmath}

\journal{Journal of Ocean Technology}

\begin{document}

\begin{frontmatter}



\title{A machine vision meta-algorithm for automated recognition of underwater objects using sidescan sonar imagery}

\author{Guillaume Labbe-Morissette}
\author{Sylvain Gauthier}
\address{CIDCO - Development Center for Ocean Mapping, Rimouski, Canada}
\address{\{guillaume.morissette,sylvain.gauthier\}@cidco.ca}

\begin{abstract}

This paper details a new method to recognize and detect underwater objects in real-time sidescan sonar data imagery streams, with case-studies of applications for underwater archeology, and ghost fishing gear retrieval. We first synthesize images from sidescan data, apply geometric and radiometric corrections, then use 2D feature detection algorithms to identify point clouds of descriptive visual microfeatures such as corners and edges in the sonar images. We then apply a clustering algorithm on the feature point clouds to group feature sets into regions of interest, reject false positives, yielding a georeferenced inventory of objects.

\end{abstract}


\end{frontmatter}

\section{Introduction}
\label{intro}

The problem of finding underwater objects is a recurring issue in many fields, such as hydrography, search and rescue (SAR), underwater archaeology, marine sciences, and many more. Unfortunately, the hostile nature of the underwater environment for human beings, the difficulty of acquiring high-quality images, along with the high mobilizing costs of SCUBA or remotely-operated solutions only makes this endeavour harder to fulfill.

The shift towards autonomous vehicles equipped with acoustic imaging technology as force multipliers during such operations brings along new problems with the multiplication of data sources, such as an exponential increase in required resources for post-processing the data. This justifies the need for real-time automation to cut costs and delays between data acquisition, interpretation and actionable results.

Therefore, we hereby propose a novel method to rapidly detect objects from sidescan sonar images to provide a real-time georeferenced catalog of objects in order to increase situational awareness. We also provide two case studies showcasing applications in different fields, namely underwater archaelogy and ocean waste management.

\section{Previous work}
\label{previouswork}

Much work has been done into the area of detecting known objects inside images using various descriptors, such as SIFT \citep{Lowe1999}, SURF \citep{Bay2008}, BRIEF \citep{Calonder2010} or ORB \citep{Rublee2011}.  Unfortunately, all of these methods require apriori knowledge of the objects to be found, and require a preliminary training stage using known data in order to adequately detect those objects in new data. 

More recently, the rise in popularity of convolutional neural network (CNN) methods have given birth to interesting detectors such as AlexNet \citep{Krizhevsky2012}, VGG \citep{vgg2015}, or \\GoogLeNet \citep{googlenet2015}, to name only a few who, while sporting very respectable classification figures, still suffer from the apriori knowledge pitfall and require non-negligeable training. 

Our algorithm aims to cater to those pitfalls for domains where exact detection is not always possible, or warranted, either because of the difficulty of accumulating enough data to reliably train CNNs, or because of the diversity of possible targets. As an example, we provide two such case-studies, one where underwater archaeologists need to track hardly categorizable ship debris and parts, and another one where ocean waste managers need to track the countless models of fishing gear abandoned or lost at sea.

\section{Methodology}
\label{method}

We have devised a 3 stage workflow from raw sensor stream data to actionable object information. In the first phase, we synthesize images from XTF data, which can be either streamed directly through a network connexion, or bundled into files by a data acquisition system. Either way, the resulting output is a corrected image suitable for automated analysis.

In a second phase, we generate feature point clouds to detect areas of interest. Since it has been demonstrated that objects in images can be expressed as a large set of smaller visual features\citep{ViolaJones2004}, we can reasonably expect dense feature clusters to appear around objects, along with a fairly substantial amount of noise. At this stage, the detection problem has therefore been reduced to a clustering and noise-rejection problem.

Consequently, in a third phase, we use a clustering algorithm on the feature cloud to find areas of higher feature density, which directly correlate with the presence of objects in the image. Computing the centroid of each feature cluster yields a well-defined and easily georeferenced region of interest (ROI) for each cluster, resulting in a catalog of geolocated objects scanned by the sonar.

\begin{center}
\includegraphics[scale=0.63]{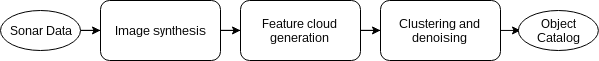} 
\\
\small{The complete data processing workflow}
\end{center}

\subsection{Image Synthesis}
\label{imagesynthesis}

Due to its fast data acquisition rate, wide area of surveying, and relatively low price point, the sidescan sonar is the current tool of choice for rapidly imaging large bodies of water \citep{blondel2009}. The sonar takes signal strength measurements, either amplitude or phase-based, through several transducer arrays known as \textit{channels}. Each channel receives sequences of vectors of quantized echo samples of the form:

\[\vec{ping}={\{sample_1,...,sample_n\}}\]

The sample count and resolution vary from model to model, but as such they can be used as row pixels by using their value as the pixel intensity, while the vectors can be stacked vertically to generate a full-size greyscale image of the channel. Typically, sidescan sonars provide at least two channels, port and  starboard.

\[
Image_{\text{n x m}} =
  \begin{bmatrix}
    ping_1 \\
    ping_2 \\
    ... \\
    ping_m \\
  \end{bmatrix} 
  =
  \begin{bmatrix}
    sample_{\text{1,1}} & ... & sample_{\text{1,n}} \\
    sample_{\text{2,1}} & ... & sample_{\text{2,n}} \\
    ... & ... & ... \\
    sample_{\text{m,1}} & ... & sample_{\text{m,n}} \\
  \end{bmatrix}   
\]

Assuming a stable surveying platform, few corrections are necessary to synthesize intelligible images suitable for automated detection. To keep preprocessing time to a minimum, we only applied slant-range correction and histogram equalization \citep{blondel2009}. 

After correction, measurements can be made in the across-track direction using the ground-range distance divided by the number of samples, and in the along-track axis using interpolated GNSS values contained in the XTF stream.

\subsubsection{Slant-range Correction}

\begin{center}
\includegraphics[scale=0.5]{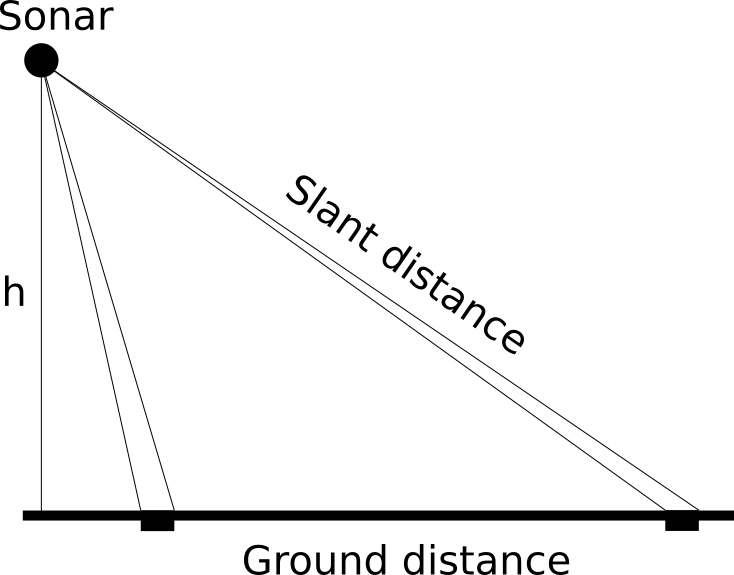} 
\\
\small{Slant-range scale distorsion.}
\end{center}

Sidescan sonar data contains noticable aberrations due to slant-range scale distorsion, which causes identically sized objects to vary in size depending on their distance from the sonar. The correction between the true distance along the ground as a function of the distance along the slant can be found through the following equation \citep{blondel2009}.

\[
Distance_\text{Ground} = \sqrt{Distance_\text{Slant}^2-h^2}
\] 

with \textit{h} being the height of the sonar taken at the nadir, and the slant distance either obtained directly through XTF data or computed as follows : 

\[
	Distance_\text{Slant}=\frac{ct_\text{twtt}}{2}
\]

using the sound speed as $c$ and $t_\text{twtt}$ the two-way travel time of the acoustic beam from the sonar to  the bottom. Should $h$ be unavailable, it can be computed using the sonar beam's tilt angle along with its roll angle (if available) through elemetary trigonometry.

\begin{center}
\includegraphics[scale=0.35]{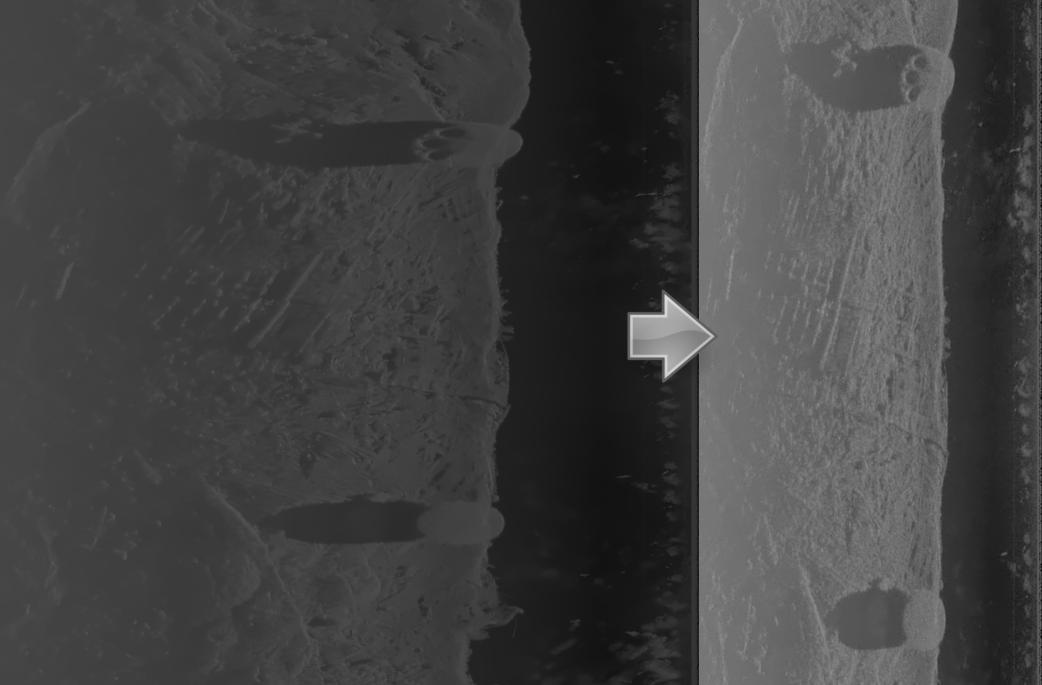} 
\\
\small{Slant-range scale correction.}
\end{center}'s ability to detect zones of interest

\subsubsection{Histogram Equalization}

The histogram equalization technique is a method to enhance the contrast in an image \citep{blondel2009}. The  process implies mapping the image's intensity histogram to another distribution, with a wider and more uniform distribution of intensity values such that the distribution covers the entire range of image values. This transform is readily available in OpenCV \citep{opencv}.

\begin{center}
\includegraphics[scale=0.35]{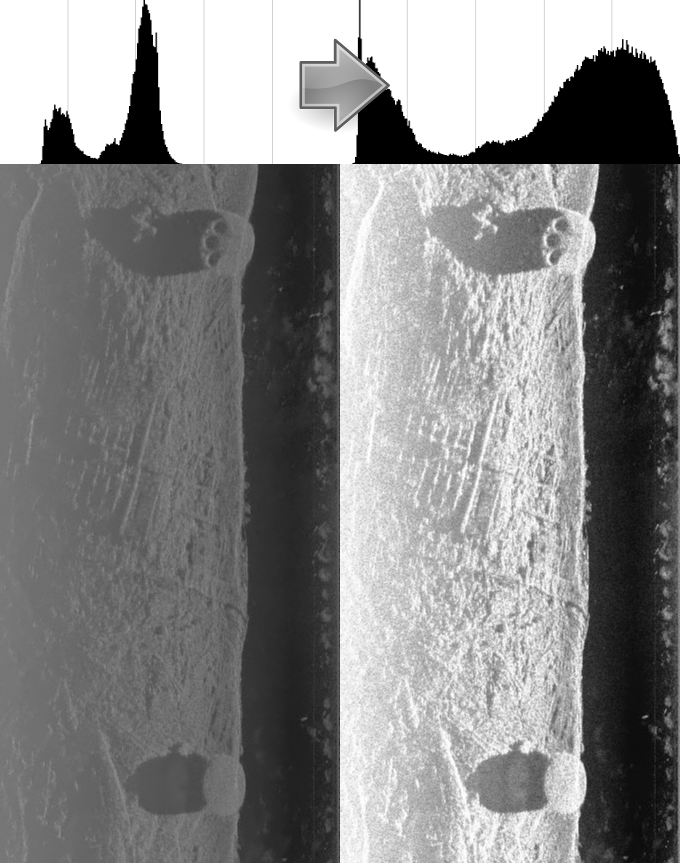} 
\\
\small{Contrast improvement through histogram equalization clearly enhances the definition and visibility of the shipwreck's boilers.}
\end{center}

\subsection{Feature Cloud Generation}
\label{featurecloudgeneration}

It has been established that the detection problem in 2D images can be reduced to finding cascades of elementary features \citep{ViolaJones2004}. As such, many classical object identification algorithms, such as SIFT \citep{Lowe1999}, SURF \citep{Bay2008} or BRIEF \citep{Calonder2010}, work by establishing sets of sensitive and specific features in order to reliably detect the target object. This has the disadvantage of requiring apriori knowledge of the objects to detect, which may not be suitable for many operational scenarios where the nature of the objects to detect might either be unknown or of a varying nature. Convolutional neural networks suffer from the same problematic apriori requirements, and as such, are not suitable at this point in the workflow. 

To overcome this issue, our approach thus consists of considering features as a measure of visual entropy,vgg2015 whose density and count are significantly higher in zones where objects are present. While most of the Harris \citep{Harris1988} or SUSAN \citep{Smith95} family of feature detectors could be suitable, we have chosen the Features from Accelerated Segment Test (FAST) \citep{rosten2005,rosten2006} and Maximally Stable External Regions (MSER) \citep{nister2008} algorithms due to their inherent speed of execution, making them especially suitable for embedding into autonomous systems.

\begin{center}
\includegraphics[scale=0.2]{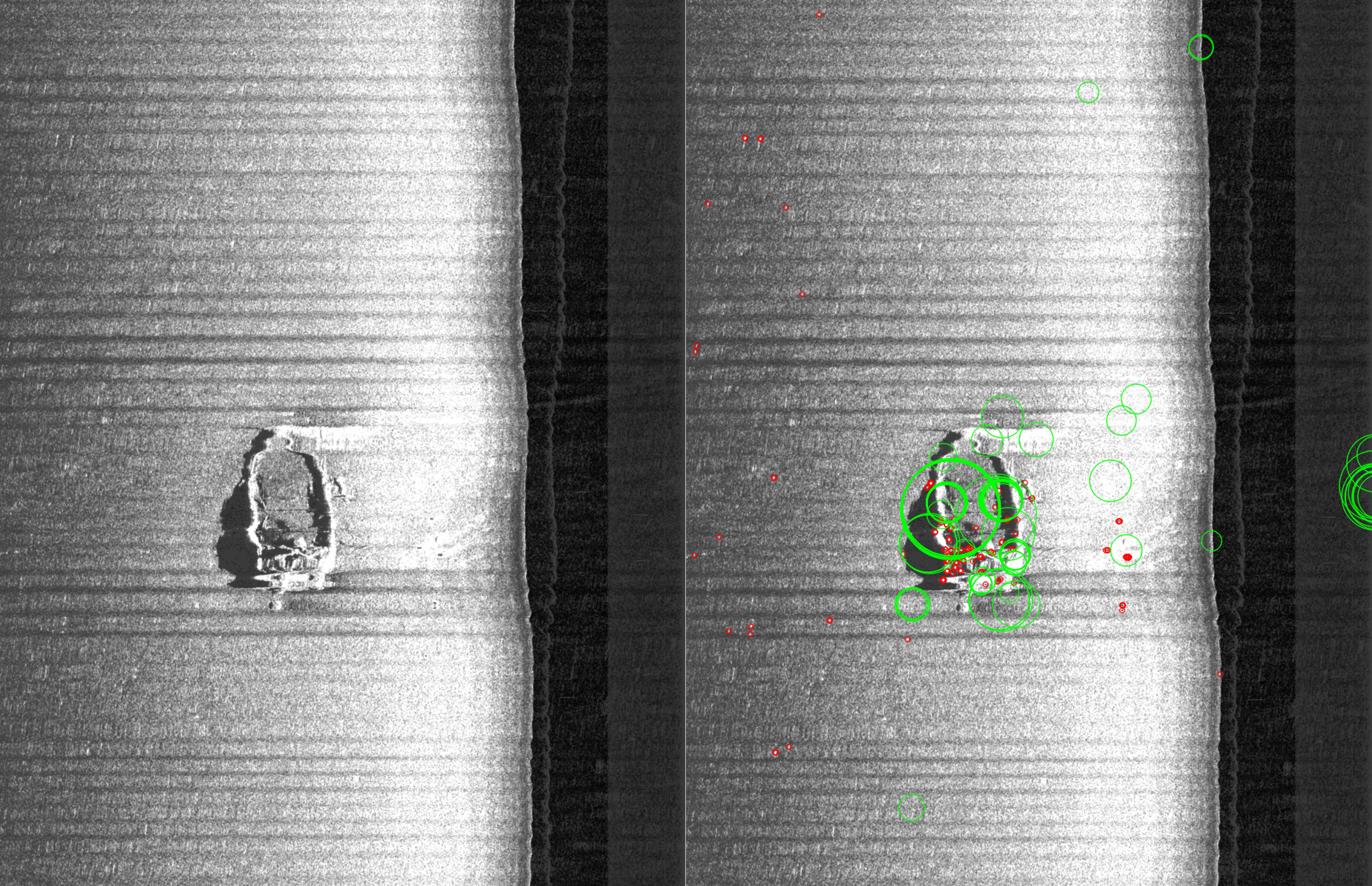} 
\\
\small{FAST (red) and MSER (green) features showing strong clusters around the region of interest containing the shipwreck and its nearby debris field.}
\end{center}

\subsection{Clustering and Denoising}
\label{clustering}

Once the previous stages of the algorithm have found enough features, objects will tend to be located in areas of higher feature density. Therefore, clustering the feature point cloud yields zones of interest inside the images. While many clustering algorithms would be adequate, DBSCAN \citep{Ester1996} provides a quick and practical solution due to its ease of use, its ability to reject noise-like features, and its support for an arbitrary number of clusters without apriori knowledge. This allows us to rapidly search our images for an arbitrary amount of clusters while discarding noise at the same time. Once the point cloud clustered, we can then compute the bounding box of the retained point features for each cluster along with a small padding value (20 pixels) to define the region of interest (ROI). 

\begin{center}
\includegraphics[scale=0.63]{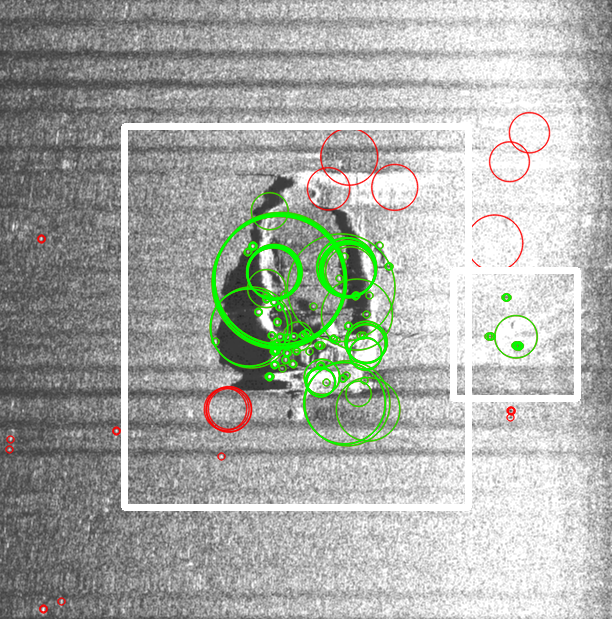} 
\\
Effective noise rejection through feature clustering, with ROI bounding boxes (white), clustered features (green), and features rejected as noise (red).
\end{center}

\section{Applications}
\label{applications}

Automatically recognizing and cataloging underwater objects yields many interesting applications in terms of surveying automation. 

\subsection{Underwater Archaeology}
\label{underwaterachaeology}

In August 2019, IRHMAS and CIDCO were involved in an archaeological campaign to find remains of the SS Germanicus, a large cargo steamer who ran aground in 1919, and of the Scotsman, one of the oldest shipwrecks in the St-Lawrence seaway, sunk in 1846, both located near Le Bic,Canada. 

\begin{center}
\includegraphics[scale=0.42]{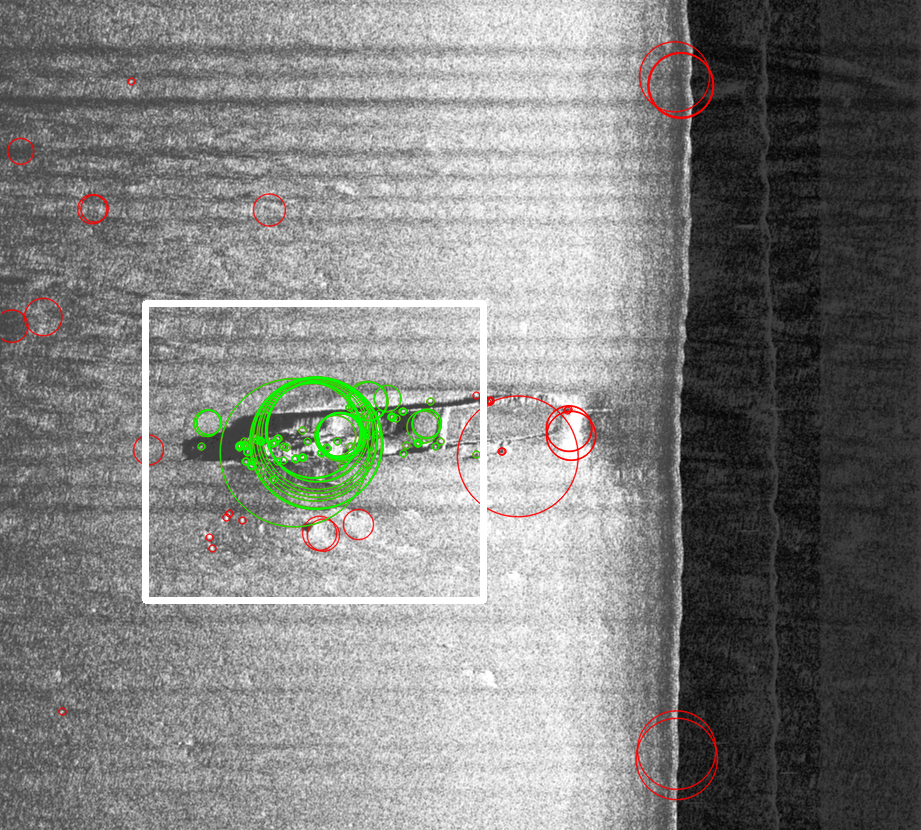} 
\\
The wreck of the SS Scotsman
\end{center}

\begin{center}
\includegraphics[scale=0.42]{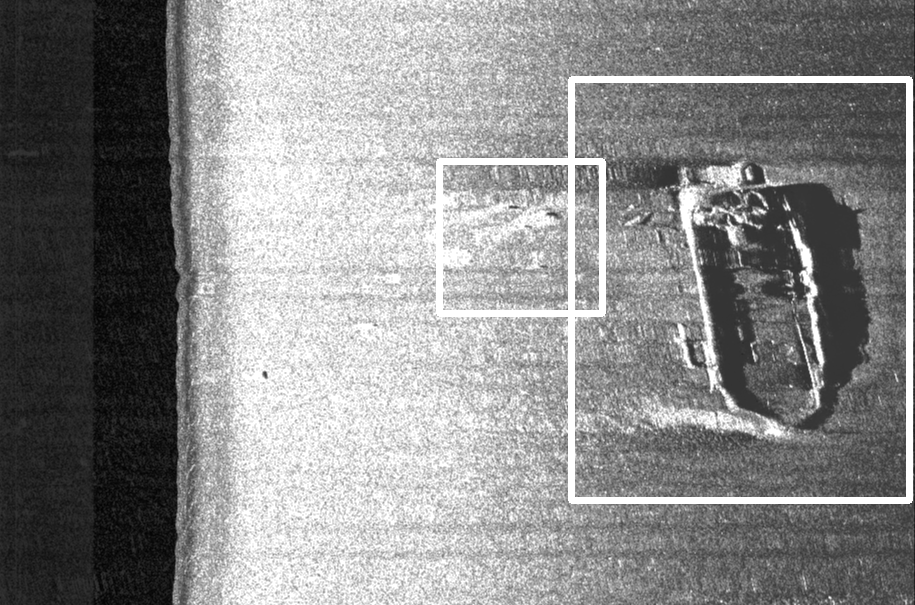} 
\\
The SS Scotsman, without feature markers, clearly showing the two detected regions of interest, the main hull and the nearby debris field.
\end{center}

The presence of large salient objects such as large hull pieces, boilers and large debris fields made this a perfect training ground for the automatic detection algorithm, which detected the large components without issues.

\begin{center}
\includegraphics[scale=0.33]{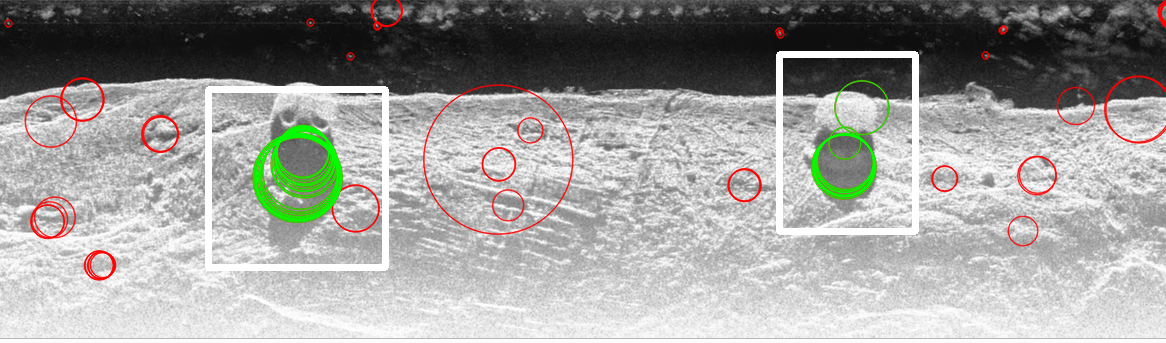} 
\\
Boilers from the SS Germanicus, surveyed with a StarFish 990F sidescan sonar.
\end{center}

\subsection{Ghost Fishing Gear}
\label{ghostfishinggear}

In 2018, CIDCO, based on advisories from Fisheries and Oceans Canada, started testing different methodologies for detecting ghost fishing gear to facilitate retrieval operations.

Ghost fishing gear is defined as fishing equipment, either abandoned, lost or thrown away, that continues to be functional in the water, therefore continuig to fullfill its function of trapping, mutilating or killing marine life in an unsupervised fashion. Encounters and entanglement with ghost fishing gear are amongst the main lethal threats for many endangered species, such as large mammals who can easily become entangled in vertical ropes suspended in the water. For example, more than 80\% of right whales in the North Atlantic (\textit{Eubalaena glacialis}) become entangled in fishing gear at least once in their lifetime \citep{knowlton2012}.

\begin{center}
\includegraphics[scale=0.4]{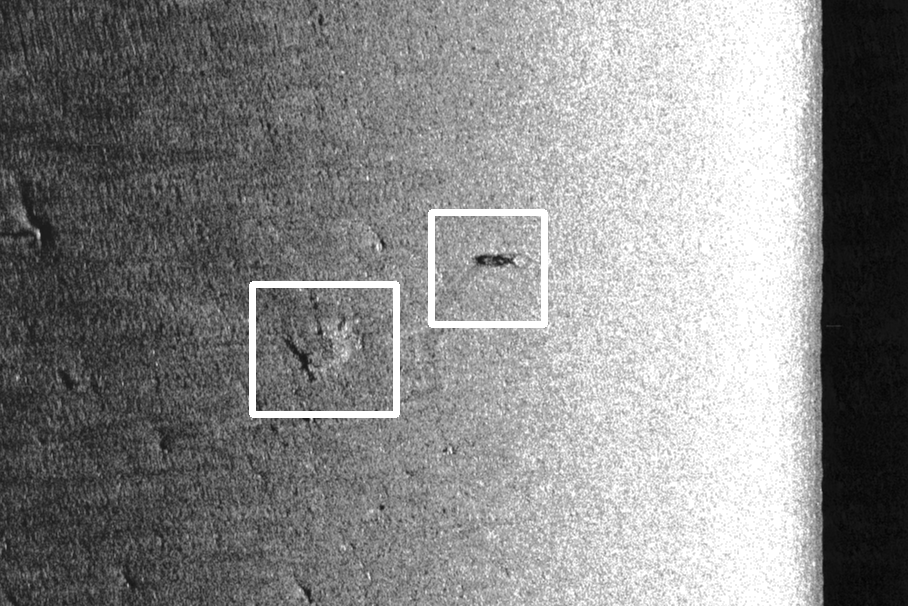} 
\\
A crab trap along with its rope, hanging in the water column. Scanned using an Edgetech 272-TD analog sidescan sonar
\end{center}

\section{Conclusion}
\label{conclusion}

The ability to automatically detect unknown objects underwater and automatically generate a georeferenced object catalog brings about new possibilities in terms of underwater surveying automation.

Our algorithm opens up interesting applications in the field of underwater archaeology, such as fully automated underwater inventories who can significantly lower the cost of surveying large areas with high-speed vessels. This allows archaeologists to do more work, and partially frees them from the dependency on hydrographers to post-process their field data. 

In the case of ghost fishing gear, field testing of our algorithm has shown efficient detection capabilities. This opens the way for major improvements in the efficiecy of ghost fishing gear retrieval processes by allowing high-speed surveying of large surfaces to adequately map retrieval zones and thus lower the cost of retrieval efforts. Further research in the matter is already under way at CIDCO.

\section{References}

\bibliographystyle{elsarticle-harv}
\bibliography{biblio.bib}




%% \bibitem must have one of the following forms:
%%   \bibitem[Jones et al.(1990)]{key}...
%%   \bibitem[Jones et al.(1990)Jones, Baker, and Williams]{key}...
%%   \bibitem[Jones et al., 1990]{key}...
%%   \bibitem[\protect\citeauthoryear{Jones, Baker, and Williams}{Jones
%%       et al.}{1990}]{key}...
%%   \bibitem[\protect\citeauthoryear{Jones et al.}{1990}]{key}...
%%   \bibitem[\protect\astroncite{Jones et al.}{1990}]{key}...
%%   \bibitem[\protect\citename{Jones et al., }1990]{key}...
%%   \harvarditem[Jones et al.]{Jones, Baker, and Williams}{1990}{key}...
%%


% \end{thebibliography}

\end{document}